\documentclass[10pt,twocolumn,a4paper]{article}

\usepackage{titling}

 \usepackage{algorithm}
 \usepackage{algpseudocode}
\usepackage{amsmath} 
\usepackage{authblk}
 \usepackage{color}
 \usepackage{comment}
 \usepackage{enumitem}
\usepackage{graphicx} 
\usepackage{latexsym}
 \usepackage{mdframed}
\usepackage{multirow} 
\usepackage{microtype}
\usepackage{subcaption} 
 \usepackage{soul} 
\usepackage{times}
\usepackage[hyphens]{url}
 \usepackage{verbatim} 
\usepackage{wrapfig}

\usepackage{algorithm}
\usepackage{algpseudocode}
\usepackage{amsmath} 
\usepackage{color} 
\usepackage{enumitem} 
\usepackage{graphicx} 
\usepackage{latexsym}
\usepackage{multirow} 
\usepackage{mdframed}
\usepackage{titling} 
\usepackage{times}
\usepackage{soul} 
\usepackage{url}
\usepackage{verbatim} 
\usepackage[hang,flushmargin]{footmisc}  

\DeclareMathOperator*{\argmax}{arg\,max}

\title{\emph{Ask, and shall you receive?}: Understanding Desire Fulfillment in Natural Language Text}

\author[1]{Snigdha Chaturvedi}
\author[2]{Dan Goldwasser}
\author[1]{Hal Daum\'{e} III}
\affil[1]{Department of Computer Science, University of Maryland, College Park}
\affil[2]{Department of Computer Science, Purdue University}

\date{}

\begin{document}
\maketitle

\begin{abstract}
The ability to comprehend desires and their fulfillment is important to Natural Language Understanding. This paper introduces the task of identifying if a desire expressed by a subject in a given short piece of text was fulfilled. We propose various unstructured and structured models that capture fulfillment cues such as the subject's emotional state and actions. Our experiments with two different datasets demonstrate the importance of understanding the narrative and discourse structure to address this task.
\end{abstract}



\section{Introduction}


Understanding expressions of desire is a fundamental aspect of understanding intentional human-behavior. 
The strong connection between desires and the ability to plan and execute appropriate actions was studied extensively in contexts of rational agent behavior \cite{georgeff1999belief}, and 
modeling human dialog interactions~\cite{grosz1986attention}.

In this paper we recognize the significant role that expressions of desire play in \textbf{natural language understanding}. Such expressions can be used to provide rationale for character behaviors when analyzing narrative text~\cite{goyal2010automatically,elson2012detecting}, extract information about human wishes~\cite{goldberg2009may}, explain positive and negative sentiment in reviews, and support automatic curation of community forums by identifying unresolved issues raised by users. 

\begin{figure}[tb]
\centering
 \includegraphics[width=\linewidth]{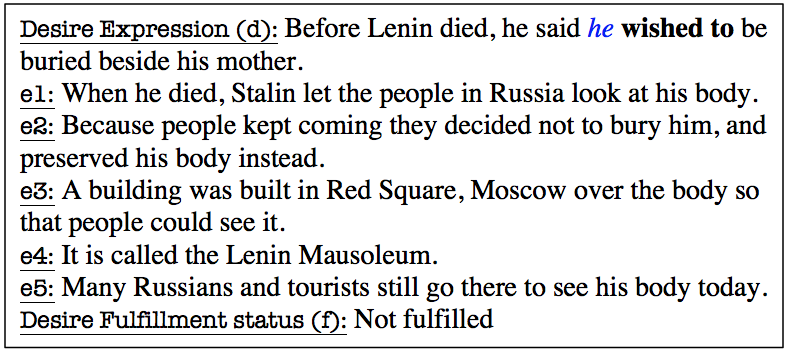}
\caption{\small{Example of a Desire Expression (d), Evidence fragments (e1$\ldots$e5) and a binary Desire Fulfillment Status (f). The Desire-subject and Desire-verb are marked in blue and bold fonts respectively in the Desire-expression.}}
  \label{fig:desireExample}
\end{figure}

We follow the intuition that at the heart of the applications mentioned above is the ability to recognize whether the expressed desire was fulfilled or not, and suggest a novel reading comprehension task: Given text, denoted as \textit{Desire-expression} (e.g., ``Before Lenin died, he said he wished to be buried beside his mother.'') containing a desire (``be buried beside his mother'') by the \textit{Desire-subject} (``he''), and the subsequent text (denoted \textit{Evidence fragments} or simply \textit{Evidences}) appearing after the Desire-expression in the paragraph, we predict if the Desire-subject was successful in fulfilling their desire. Fig.~\ref{fig:desireExample} illustrates our setting.

Similar to many other natural language understanding tasks~\cite{dagan2010recognizing,MCTest,berant-EtAl:2014:EMNLP2014}, performance is evaluated using 
prediction accuracy.  
However, unlike tasks such as text categorization or sentiment classification which rely on lexical information, 
understanding desire fulfillment requires complex inferences connecting expression of desire, actions affecting the Desire-subject, and the extent to which these actions contribute to fulfilling the subject's goals. For example, in Fig.~\ref{fig:desireExample} the action of `preserving' Lenin's body led to non-fulfillment of his desire.

We address these complexities by representing the narrative flow of  \textit{Evidence} fragments, and assessing if the events (and emotional states) mentioned in this flow contribute to (or provide indication of) fulfilling the desire expressed in the preceding Desire-expression. Following previous work on narrative representation~\cite{chambers2008unsupervised}, we track the events and states associated with the narrative's central character (the \textit{Desire-subject}). 

While this representation captures important properties required by the desire-fulfillment prediction task, such as the actions taken by the Desire-subject, it does not provide us with an indication about the outcome of these actions. Recent attempts to support supervised learning of such detailed narrative structures by annotating data~\cite{L12-1515}, result in highly complex structures even for restricted domains. Instead we model this information by associating a state, indicating if the outcome of an action (or the mention of an emotional state) provides evidence for making progress towards achieving the desired goal.  We model the transitions between states as a latent sequence model, and use it to predict if the value of the final latent state in this sequence is indicative of a positive or negative prediction for our task. 

We demonstrate the strength of our approach by comparing it against two strong baselines. First, we demonstrate the importance of analyzing the complete text by comparing with a textual-entailment based model that analyzes individual Evidence fragments independently.
We then compare our latent structured model, which incorporates the narrative structure with an unstructured model, and show improvements in prediction performance. 
Our key contributions are: 
\begin{itemize}[leftmargin=*,noitemsep,nolistsep]
\itemsep0em 
\item {We introduce the problem of understanding desire fulfillment,  annotate and release two datasets for further research on this problem.}
\item {We present a latent structured model for this task, incorporating the narrative structure of the text, and propose relevant features that incorporate world knowledge. }
\item {Empirically demonstrate that such a model outperforms competitive baselines.}

\end{itemize}

\subsection{Problem Setting}
Our problem consists of instances of short texts (called \emph{Desire-expressions}), which were collected in a manner so that each consists of an indication of a desire (characterized using a \emph{Desire-verb}) by a \emph{Desire-subject(s)}. The Desire-verb is identified by the following verb phrases: `wanted to', `wished to' or `hoped to'~\footnote{We chose to use these three phrases for data collection. However, one can include other expressions of desire if needed. We plan to include that in future work.}. The three Desire-verbs were identified using lexical matches while the Desire-subject(s) was marked manually. Each \emph{Desire-expression} is followed by five or fewer pieces of \emph{Evidence fragment}s (or simply \emph{Evidence}s). The Desire-expression and the Evidences (in order) consist of individual sentences that appeared contiguously in a paragraph. We address the binary classification task of predicting the \emph{Desire Fulfillment} status, i.e. whether the indicated desire was fulfilled in the text, given the Evidences and the Desire-expression with Desire-verb and Subject identified.  Fig.~\ref{fig:desireExample} shows an example of the problem.

\section{Inference Models for Understanding Desire Fulfillment in Narrative Text}

\label{classification}
In this section we present three textual inference approaches, each following different assumptions when approaching the desire-fulfillment task, thus allowing a principled discussion about which aspects of the narrative text should be modeled. 

Our first approach assumes the indication of desire fulfillment will be contained in a single Evidence fragment. We test this assumption by adapting the well-known Textual Entailment task to our settings, by generating entailment candidates from Desire-expression and Evidence fragments. 

Our second approach assumes the decision depends on the Evidence text as a whole, rather than on a single Evidence fragment. We test this assumption by representing relevant information extracted from the entire Evidence text. This representation (depicted in Fig.~\ref{fig:featureFramework}) connects the central character in the narrative, the Desire-subject, with their actions and emotional states exhibited in the Evidence text. This representation is then used for feature extraction when training a binary classifier for the desire-fulfillment task.

Our final model provides a stronger structure for the actions and emotional states expressed in the Evidence text. The model treats individual Evidence fragments as parts of a plan carried out by the Desire-subject to achieve the desired goal, and makes judgments about the contribution of each step towards achieving the desired goal.
 
\subsection{Textual Entailment (TE) Model}
\label{EOP}
 
Recognizing Textual Entailment (RTE) is the task of recognizing the existence of an entailment relationship between two text fragments~\cite{dagan2010recognizing}. 
From this perspective, a textual entailment based method might be a natural way to address the desire fulfillment task. 
RTE systems often rely on aligning the entities appearing in the text fragments. Hence we reduce the desire fulfillment task into several RTE instances consisting of text-hypothesis pairs, by pairing the Desire-expression (hypothesis) with each of the Evidence fragments (text) in that example. 
However, we ``normalized'' the Desire-expression, so that it would be directly applicable for the RTE task. 
For example, the Desire-expression, ``One day \emph{Jerry} \textbf{wanted to} paint his barn.", gets converted to ``Jerry painted his barn.''.
This process followed several steps: 
\vspace{-1pt}
 \begin{itemize}[noitemsep,leftmargin=*,nolistsep,topsep=0pt]
 \itemsep0em 
\item {If the Desire-subject is pronominal, replace it with the  appropriate named entity when possible (we used the Stanford CoreNLP coreference resolution system)~\cite{corenlp}.}
\item {Ignore the content of the Desire-expression appearing before the Desire-subject. }
\item {Remove the clause containing the Desire-verb (`wanted to', `wished to' etc.), and convert the succeeding verb to its past tense.}
\end{itemize} 
\vspace{-1pt}

The desire was considered `fulfilled' if the RTE model predicted \emph{entailment} for at least one of the text-hypothesis pairs of the example. E.g., the model could infer that the normalized Desire-expression example mentioned above, would be entailed by the following Evidence fragment- ``It took Jerry six days to paint his barn that way.'' and hence it would conclude that the desire was fulfilled. 
Table~\ref{table:TEResults} shows the performance of BIUTEE~\cite{BIUTEE,EOP}, an RTE system, on the two datasets (see Sec.~\ref{datasets}) used in our experiments\footnote{We also tested the TE model by using the default setting, optimized for the RTE task, however it performed very poorly.}. 
Our results show that the RTE Model performs better with normalization. We use this model (with normalization) as a baseline in Sec.~\ref{experiments}.

\begin{table}
\centering
\footnotesize{
\begin{tabular}{|c|c|r|r|r|}\hline
\textbf{Data}   & \textbf{Normalized?}  & \textbf{P}    & \textbf{R}    & \textbf{F}  \\\hline
 MC             & No                    & 59.38         & 24.68         & 34.86 \\\cline{2-5}
 Test           & Yes                   & 76.09         & 45.45         & \textbf{56.91} \\\cline{1-5}
Simple          & No                    & 50.00         & 2.22          & 4.26  \\\cline{2-5}
Wiki            & Yes                   & 37.04         & 8.89          & \textbf{14.34} \\\cline{1-5}
\end{tabular}
}
\vspace{0.05in}
\caption{\small{Normalizing the Desire-expression helps the TE model.}}
\label{table:TEResults}
\end{table}

\subsection{Unstructured Model}
\label{flatmodels}
The Textual Entailment model described above assumes that the Desire-expression would be entailed by one of the individual Evidences.  
This assumption might not hold in all cases. Firstly, the indication of desire fulfillment (or its negation) can be subtle and expressed using indirect cues. 
More commonly, multiple Evidence fragments can collectively provide the cues needed to identify desire fulfillment. 
This suggests a need to treat the entire text as a whole when identifying cues about desire fulfillment. 

We begin by identifying the Desire-subject and the desire expressed (using `focal-word' described in Sec.~\ref{features}) in the Desire-expression. Thereafter, we design several semantic features to model coreferent mentions of the Desire-subject, actions taken (and respective semantic-roles of the Desire-subject), and emotional state of the Desire-subject in the Evidences. We enhance this representation using several knowledge resources identifying word connotations~\cite{connotation} and relations. Fig.~\ref{fig:featureFramework} presents a visual representation of this process and Sec.~\ref{features} presents further details. 

Based on these features, extracted from the collection of all Evidences instead of individual Evidence fragments, we train supervised binary classifiers (\textit{Unstructured models}).

\begin{figure}[tb]
\centering
\includegraphics[width=0.75\linewidth]{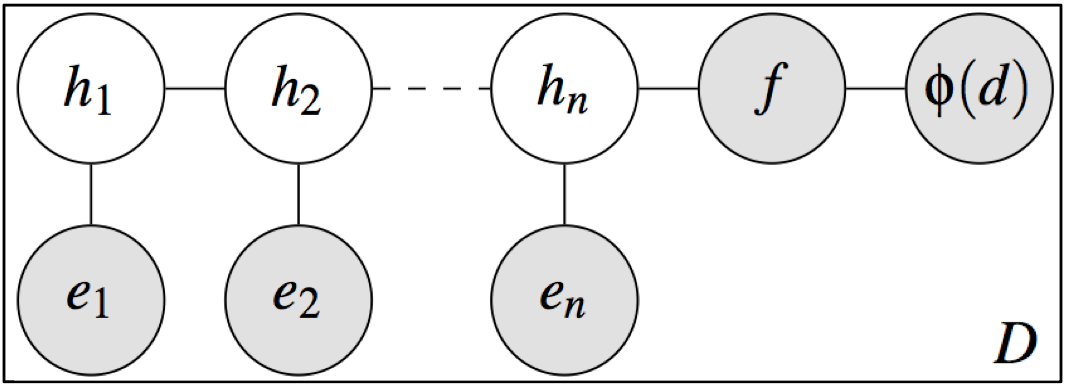}
\caption{\small{Structured model (LSNM) Diagram. Evidence $e_i$, Desire Fulfillment, $f$, and Structure-independent features, $\boldsymbol{\phi(d)}$, are observed, States, $h_i$, are hidden.}}
\label{fig:linear_diagram}
\end{figure}

\subsection{Latent Structure Narrative Model (LSNM)}
\label{lcmm}


The Unstructured Model described above captures nuanced indications of desire-fulfillment, by associating the Desire-subject with actions, events and mental states. However, it ignores the narrative structure as it fails to model the `flow of events' depicted in the transition between the Evidences.

Our principal hypothesis is that the input text presents a story. The events in the story describe the evolving attempts of the story's main character (the Desire-subject) to fulfill its desire. Therefore, it is essential to understand the flow of the story to make better judgments about its outcome. 

 We propose to model the evolution of the narrative using latent variables. We associate a latent state (denoted $h_j$), with each Evidence fragment (denoted~$e_j$). The latent states take discrete values (out of $H$ possible values, where $H$ is a parameter to the model), which abstractly represent various degrees of optimism or pessimism with respect to fulfillment, $f$ of the desire expressed in the Desire-expression, $d$. These latent states are arranged sequentially, in the order of occurrence of the corresponding Evidence fragments, and hence capture the evolution of the story (see Fig.~\ref{fig:linear_diagram}).  
 
The linear process assumed by our model can be summarized as: The model starts by predicting the latent state, $h_0$, based on the first Evidence, $e_0$. Thereafter, depending on the current latent state, and the content of the following Evidence fragment, the model transitions to another latent state. This process is repeated until all the Evidence fragments are associated with a latent state. We formulate the transition between narrative states as sequence prediction. We associate a set of \emph{Content} features with each latent state, and \emph{Evolution} features with the transitions between states. 

Note that the desire fulfillment status, $f$, is viewed as an outcome of this inference process and is modeled as the last step of this chain using a discriminative classifier which makes its prediction based on the final latent state and a \emph{Structure-independent} feature set, $\boldsymbol{\phi(d)}$. This feature set can be handcrafted to include information that could not be modeled by the latent states, such as long-range dependencies, and other cumulative features based on the Desire-expression, $d$, and the Evidence fragments, $e_j$s.

We quantify these predictions using a linear model which depends on the various features, $\boldsymbol{\phi}$, and corresponding weights, $\mathbf{w}$. Using the Viterbi algorithm we can compute the score associated with the optimal state sequence, for a given input story as:
 \vspace{-0.29cm}
 \begin{equation}
 \text{score} = \max_\mathbf{h} [\mathbf{w} \cdot \boldsymbol{\phi}(\mathbf{e}, d, f, \mathbf{h})]
\label{eqn:svm1}
\end{equation}
 
 
\vspace{-0.15in}
\begin{algorithm}[ht!]
\caption{Training algorithm for LSNM}
\begin{algorithmic}[1]
\State\textbf{Input:} Labeled set $\{(d,\mathbf{e},f)_{i}$ $\forall i \in \{1 \ldots D\}\}$; and $T$: number of iterations
\State\textbf{Output:} Weights $\mathbf{w}$
\State\textbf{Initialization:} Initialize $\mathbf{w}$ randomly
\For {$t:1$ to $T$}
\State $\hat{\mathbf{h}}_i = \argmax_{\mathbf{h}_i} [\mathbf{w_{t-1}} \cdot \boldsymbol{\phi}(\mathbf{e}_i, d_i, f, \mathbf{h}_i)]$ such that $f=f_i \forall i \in \{1 \ldots D\}$ 
\State $\mathbf{w_{t}}$ = StructuredPerceptron($\{(d, \mathbf{e}, \hat{\mathbf{h}}, f)_i\}$)
\EndFor
\end{algorithmic}
\label{fig:trainingAlgo}
\end{algorithm}
\vspace{-0.15in}

\subsubsection{Learning and Inference}
During training, we maximize the cumulative scores of all data instances using an iterative process (Alg.~\ref{fig:trainingAlgo}). Each iteration of this algorithm consists of two steps. In the first step, for every instance, it uses Viterbi algorithm (and weights from previous iteration, $\mathbf{w_{t-1}}$) to find the highest scoring latent state sequence, $\mathbf{h}$, that agrees with the provided label (the fulfillment state), $f$. In the following step, it uses the 
state sequence determined above to get refined weights for the $t^{th}$ iteration, $\mathbf{w_{t}}$, using structured perceptron~\cite{Collins:2002}.
The algorithm is similar to an EM algorithm with `hard' assignments albeit with a different objective. 
While testing, we use the learned weights and Viterbi decoding to compute the fulfillment state and the best scoring 
state sequence.
Our approach is related to latent structured perceptron though we only use the last state (and structure-independent features) for prediction. 

\section{Features}
\label{features}
\begin{table*}
\centering
\footnotesize{
\begin{tabular}{|p{0.175\linewidth}|p{0.02\linewidth}|p{0.73\linewidth}|}\hline
\textbf{Feature Type}     & \textbf{Id}       & \textbf{Definition}\\\hline
Entailment       & F1       & \emph{TEPrediction}: Binary prediction of the Textual Entailment model~\cite{BIUTEE}.\\\hline
Discourse        & F2, F3   & \emph{ButPresent, SoPresent}: Binary features indicating if a `but' or `so' (respectively) followed the Desire-verb (`wanted to', `wished to' etc.) in the Desire-expression.\\\hline
Focal Word       & F4, F5, F6       & \emph{focal count, focal syn and focal ant count:} Count of occurrences of the focal word(s), their WordNet~\cite{wordnet} synonyms and antonyms (respectively) in the Evidence. Occurrences of synonyms or antonyms were identified only when they had the same POS tag as the focal word(s). \\\cline{2-3}
                 & F7       & \emph{focal+syn count:} Sum of F4 and F5\\\cline{2-3}
                 & F8       & \emph{focal lemm count:} Count of occurrences of lemmatized forms of the focal word(s) in the Evidence.\\\hline
Desire-subject mentions & F9& \emph{sub count:} Count of all mentions (direct and co-referent) of the Desire-subject in the Evidence.\\\hline
Emotional State  & F10, F11 &  \emph{+adj, -adj count:} Counts of occurrences of `positive' and `negative' adjectives (respectively) modifying the direct and co-referent mentions of the Desire-subject in the Evidence.\\\hline
Action           & F12, F13  &  \emph{+Agent, -Agent count:} Number of times the connotation of verbs appearing in the Evidence agreed with and disagreed with (respectively) that of the intended action.\\\cline{2-3}
                 & F14, F15  & \emph{+Patient, -Patient count:} Count of occurrences of `positive' and `negative' verbs (respectively) in the Evidence which had the Desire-subject as the patient.\\\hline
Sustenance       &F16, F17 & \emph{isConforming, isDissenting:} Binary features indicating if the Evidence starts with a conforming or dissenting phrase (respectively). See Table~\ref{table:sustenanceWordLists} for example phrases.\\\hline
\end{tabular}
}
\caption{\small{Feature definitions (Sec.~\ref{features}). F1-F3 are extracted for each example while F4-F17 are extracted for each evidence.}}
\label{table:AllFeatures}
\end{table*} 

\begin{figure}[tb]
\centering
 \includegraphics[width=0.9\linewidth]{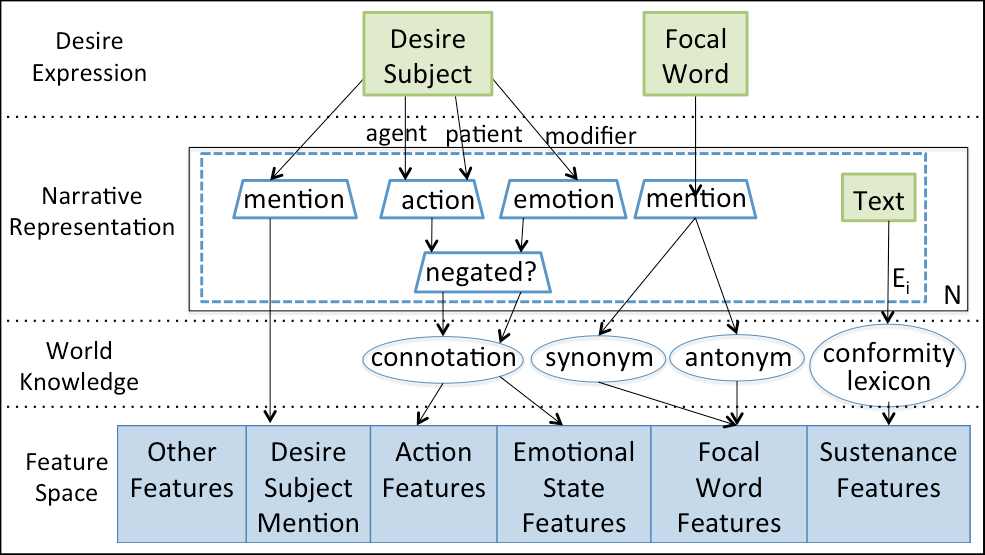}
  \caption{\small{Framework for feature extraction for an example. $E_i$ refers to the $i^{th}$ evidence out of a total of $N$ evidences.
}}
  \label{fig:featureFramework}
  \vspace{0.02in}
\end{figure}


We now describe our features and how they are used by the models.  
Table~\ref{table:AllFeatures} defines our features and Fig.~\ref{fig:featureFramework} describes their extraction for an example. They capture different semantic aspects of the desire-expression and evidences, such as entities, their actions and connotations, and their emotive states using   lexical resources like Connotation Lexicon~\cite{connotation}, WordNet and our lexicon of conforming and dissenting phrases. Before extracting features, we pre-processed the text~\footnote{We obtained pos tags, dependency parses, and resolved co-references using Stanford CoreNLP~\cite{corenlp}.} and extracted all adjectives and verbs (with their negation statuses and connotations) associated with the Desire-subject using dependency-parsing based rules. 

\begin{figure}[tb]
 \includegraphics[width=\linewidth]{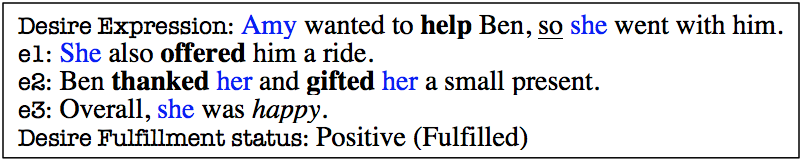}
  \caption{\small{Artificial example indicating feature utility. The Desire-subject mentions are marked in blue, actions in bold and emotions in italics. Discourse feature is underlined.}}
  \label{fig:artificialEg}
\end{figure}

\noindent \textbf{1. Entailment (F1):} This feature simply incorporates the \emph{output of the Textual Entailment model}. 

\noindent \textbf{2. Discourse (F2-F3):} These features aim to identify \emph{indications of obstacles or progress} of desire fulfillment in the Desire-expression itself, based on discourse connectives. E.g. `so' (underlined) in the Desire-expression in Fig.~\ref{fig:artificialEg} indicates progress of desire fulfillment.


\noindent \textbf{3. Focal words (F4-F8):} 
These features identify the \emph{word(s) most closely related} to the desire, and look for their presence in the Evidences. 
We define a \textit{focal word} as the clausal complement of the Desire-verb (`wanted to', `hoped to', `wished to'). If the clausal complement is a verb, the focal word is its past tense form. e.g., the focal word in the Desire expression in Fig.~\ref{fig:artificialEg} is `helped'. A focal word is not simply the verb following the Desire-verb: e.g. in the Desire-expression in Fig.~\ref{fig:desireExample}, the causal complement of `wished' is `buried'. 
We then define features counting occurrences of the identified \emph{focal words} and their WordNet \emph{synonyms and antonyms} in each of the Evidences.


\noindent \textbf{4. Desire-subject mentions (F9):} This feature looks for mentions of Desire-subject in the Evidences 
assuming that a \emph{lack of mentions} of the Subject might indicate absence of instances of their taking actions needed to fulfill the desire. 


\noindent \textbf{5. Emotional State (F10-F11):} Signals about the fulfillment status could also emanate from the \emph{emotional state of the Subject}. A happy or content Desire-subject can be indicative of a fulfilled desire
(e.g. in Evidence e3 in Fig.~\ref{fig:artificialEg}), and vice versa. We quantify the emotional state of the Subject(s) using connotations of the adjectives modifying their mentions. 

\noindent \textbf{6. Action features (F12-F15):} These features 
analyze the \emph{intended} action and the actions taken by various entities. We first identify the \emph{intended action} - the verb immediately following the Desire-verb in the Desire Expression. e.g., in Fig.~\ref{fig:artificialEg}
the intended action is to `help'. Thereafter, we design features that capture the connotative agreement between the intended action and the actions taken by the Desire-subject(s) in the Evidences. We also include features that describe connotations of actions (verbs) affecting the Desire-subject(s). E.g. in e1 of Fig.~\ref{fig:artificialEg}, the action by the Desire-subject (marked in blue), `offered', is in connotative agreement with the intended action, `help' (both have positive connotations according to~\cite{connotation}). Also, the actions affecting the subject (`thanked', `gifted') have positive connotations indicating desire fulfillment. 

\begin{table}
\centering
{\footnotesize
\begin{tabular}{|l|l|}\hline
Type & Phrases\\\hline
Conforming  & \emph{in other words, for example, consequently,}\\
            & \emph{apparently because, hence, especially since}\\\hline
Dissenting  & \emph{although, but, by contrast, conversely,}\\
            & \emph{even though, however, instead, meanwhile}\\ \hline
\end{tabular}}
\vspace{0.05in}
\caption{\small{Some examples of conforming and dissenting phrases.}}
\label{table:sustenanceWordLists}
\end{table} 




\noindent \textbf{7. Sustenance Features (F16-F17):} 
LSNM uses a chain of latent states to abstractly represent the content of the Evidences with respect to Desire fulfillment Status. At any point in the chain, the model has an expectation of the fulfillment status. The sustenance features \emph{indicate if the expectation should intensify, remain the same or be reversed by the incoming Evidence fragment}. This is achieved by designing features indicating if the Evidence  fragment starts with a `conforming' or a `dissenting' phrase. E.g. e3 in Fig.~\ref{fig:artificialEg} starts with a conforming phrase, `Overall', indicating that the fulfillment status expectation (positive in e2) should not change. Table~\ref{table:sustenanceWordLists} presents some examples of the two categories. These phrases were chosen using various discourse senses mentioned in \cite{pdtbManual}. The complete list is available on the first author's webpage.

\subsection{Unstructured Models}
\label{feat_unstructuredModels}
For the unstructured models, we directly used the Entailment and Discourse features (F1 to F3 in Table~\ref{table:AllFeatures}). For features F4 to F15, we summed their values across all Evidences of an instance. This ensured a constant size of the feature set in spite of variable number of Evidence fragments per instance.


\subsection{Latent Structure Narrative Model}
\label{feat_structuredModels}
Our Structured model requires three types of features: (a) Content features that help the model assign latent states to Evidence fragments based on their content, (b) Evolution features that help in modeling the evolution of the story expressed by the Evidence fragments (c) Structure Independent features used while making the final prediction. 

\noindent \textbf{Content features:} These features depend on the latent state of the model, $h_j$, and the content of the corresponding Evidence, $e_j$ (expressed using features F4 to F15 in Table~\ref{table:AllFeatures}). 
\begin{enumerate}[noitemsep,topsep=0pt]
\itemsep0em 
\item $\phi(h_j,e_j) =$ $\alpha$ if the current state is $h_j$; $0$ otherwise where  $\alpha \in$ F4 to F15
\end{enumerate}

\noindent \textbf{Evolution features:} These features depend on the current and previous latent states, $h_j$ and $h_{j-1}$ and/or the current Evidence fragment, $e_j$:
\begin{enumerate}[noitemsep,topsep=0pt]
\itemsep0em 
\item $\phi(h_{j-1}, h_j) =$ $1$ if previous state is $h_{j-1}$ and current state is $h_{j}$; $0$ otherwise.
\item $\phi(h_{j-1}, h_j, e_j) =$ $\alpha$ if previous state is $h_{j-1}$, current state is $h_{j}$; $0$ otherwise  where  $\alpha \in$ F16 and F17
\item $\phi(h_{0}) =$ $1$ if start state is $h_{0}$; $0$ otherwise.
\end{enumerate}

\noindent \textbf{Structure Independent features $\boldsymbol{\phi(d)}$:} 
This feature set is exactly same as that used by the Unstructured models.

\section{Datasets}

\label{datasets}

We have used two real-world datasets for our experiments: MCTest and SimpleWiki consisting of 174 and 1004 manually annotated instances respectively. Both the datasets (available on the first author's webpage) were collected and annotated in a similar fashion. 

\textbf{Collection and annotation:} The MCTest data originated from the Machine Comprehension Test dataset~\cite{MCTest} which contained of a set of 660 stories and associated questions. 
The vocabulary and concepts are limited to the extent that the stories would be understandable by 7 year olds. We discard the questions and only consider the free text of the stories.

The SimpleWiki dataset was created from the textual content of an October, 2014~\footnote{http://dumps.wikimedia.org/simplewiki/} dump of the Simple English Wikipedia. 
We discarded all lists, tables and titles in the wiki pages. We chose Simple English Wikipedia instead of Wikipedia articles to limit the complexity of the vocabulary and world knowledge required to comprehend the content thus making the task simpler and manageable.


The Desire-subject(s) and the Desire Fulfillment Status were manually annotated on CrowdFlower~\footnote{http://www.crowdflower.com/}. 
Each instance was annotated by 3 or more annotators as determined by CrowdFlower using expected annotation accuracy. 
Annotators were also required to demonstrate proficiency on an initial set of 5 test instances. To avoid 
annotator fatigue, each annotator was presented only 3 instances 
per session. The mean CrowdFlower confidence (inter-annotator agreement weighted by their trust scores) of the annotations was 0.92. 



\textbf{Training and Test Sets: } The SimpleWiki and MCTest data consisted of about 1000 and 175 instances, 20\% of which was held-out as test sets. In the test sets of SimpleWiki and MCTest, 28\% and 56\% of the data belonged to the positive (desire fulfilled) class respectively.


\section{Empiricial Evaluation}
\label{experiments}
\label{mainResults}

\begin{table}
\centering
\footnotesize{
\begin{tabular}{|c|c|c|r|r|r|}\hline
\textbf{Data} & \textbf{Model type} & \textbf{Name} & \textbf{P} & \textbf{R} & \textbf{F} \\\hline
        & Bag-Of-Words                      & BoW           &  41.2 & 50.0  & 45.2          \\\cline{2-6}
        & Textual Entailment                & TE            &  76.1 & 45.4  & 56.9          \\\cline{2-6}
 MC     & \multirow{ 2}{*}{Unstructured}    & LR            &  70.6 & 63.2  & 66.7          \\\cline{3-6}
Test    &                                   & DT            &  71.4 & 52.6  & 60.6          \\\cline{2-6}
        & Structured                        & \textbf{LSNM} &  69.6	& 84.2  & \textbf{74.4} \\\hline\hline
        & Bag-Of-Words                      & BoW           &  28.2 & 20.0  & 23.4          \\\cline{2-6}
        & Textual Entailment                & TE            &  37.0 & 8.9   & 14.3          \\\cline{2-6}
Simple  & \multirow{ 2}{*}{Unstructured}    & LR            &  50.0 & 8.9   & 15.2          \\\cline{3-6}
Wiki    &                                   & DT            &  42.9 & 5.4   & 9.5           \\\cline{2-6}
        & Structured                        & \textbf{LSNM} &  37.5 & 21.3  & \textbf{27.1}   \\\hline
\end{tabular}
}
\vspace{0.05in}
\caption{\small{Test set performances. Our structured model, LSNM, outperforms the unstructured, TE and BoW models.}}
\label{table:AllResults}
 \end{table}

For evaluation, we compared test set performances using F1 score of the positive (desire fulfilled) class. We also included a simple Logistic Regression baseline based on Bag-of-Words (BoW) features. Table~\ref{table:AllResults} reports the performances of these models. For training the unstructured model, we experimented with different algorithms and show the results for the best two models: LR (Logistic Regression) and DT (Decision Trees). We report median performance values over $100$ random restarts of our model since its performance depends on the initialization of the weights. Also, our model requires the number of latent states, $H$, as input which was set to be $2$ and $15$ 
for the MCTest and SimpleWiki datasets respectively using cross-validation. The  difference in optimal $H$ values (and F1 scores) for the two datasets could be attributed to the difference in complexity of the language and concepts used in them. The MCTest dataset consists of children stories, focusing on simple concepts and goals (e.g., `wanting to go skating') and their fulfillment is indicated explicitly, in simple and focused language (e.g.,  ‘They went to the skating rink together.’). On the other hand, SimpleWiki describes real-life desires (e.g., `wanting to conquer a country'), which require sophisticated planning over multiple steps, which may provide only indirect indication of the desire fulfillment status. This added complexity resulted in a harder classification problem, and increased the complexity of inference over several latent states.


The table shows that LSNM outperforms the unstructured models indicating the benefit of modeling narrative structure. Also, the unstructured models perform better than the TE model emphasizing the need for simultaneous analysis all of the Evidence text.
We obtained similar results during cross validation. For instance, the TE, unstructured models (best) and LSNM yielded F1 scores of $56.9$, $67.9$ and $70.2$ respectively on the MCTest data. This shows that modeling the narrative presented by the Evidences results in better prediction of the desire fulfillment status.

\section{Related Work}
\label{related}

Expressions of desires and wishes have attracted psycholinguists~\cite{Shatz1983} and linguists~\cite{barak2013acquisition} alike. \cite{goldberg2009may} detect wishes from text. Analyzing desires adds a new dimension to more general tasks like opinion mining~\cite{Pang2007} where the manufacturers and advertisers want to discover users' desires or needs from online reviews etc. Another use-case would be in resolving issues for community forum users. For instance, the number of posts in Massive Open Online Courses forums often overwhelm the instructional staff~\cite{chaturvedi-goldwasser-daumeiii:2014:P14-1}. Identifying posts containing unresolved issues can help focus the efforts of the instructional staff.

Our problem is related to Machine Comprehension~\cite{MCTest}. However, unlike most systems, designed for understanding large textual collections (\emph{macro-reading})~\cite{Etzioni2006,Carlson2010,Fader2011}, this work focuses on \emph{Micro-reading}, understanding short pieces of text. \cite{berant-EtAl:2014:EMNLP2014} also address micro-reading but with a different goal --  answering domain-specific questions about entities in a paragraph.

Our task is also related to Recognizing Textual Entailment (RTE)~\cite{dagan2010recognizing,Dagan2005}. However, we show that solving it additionally requires modeling the narrative structure of the text.

There have been several attempts at modeling narrative structures which include narrative schemas~\cite{chambers2009unsupervised,chambers2008unsupervised}, plot units~\cite{lehnert1981} and Story Intention Graphs~\cite{L12-1515}. Previous work has also studied connotations and word effects on narrative modeling \cite{connotation,goyal2010automatically}. 
Our approach is closely related to these methods. While focusing on a specific classification task, our structured model and features, share similar motivation. 

The AI task of recognizing plans of characters in a narrative viewing them as intentional agents~\cite{Mueller:2007,Wilensky:1978,Mani:2013} is also relevant. However, the focused nature of our task lets us employ latent variables to model the transitions between expectations and plans.
 
 Latent structured models have been used previously for solving various problems in computer vision and NLP~\cite{Tackstrom:2011,Yessenalina:2010,Felzenszwalb:2008} though their problem settings and goals are different.

\section{Conclusion}
\label{conclusion}

In this paper we have addressed the novel task of analyzing small pieces of text containing expression of a desire to identify if the desire was fulfilled in the given text. For solving this problem, we adopt three approaches based on different assumptions. We first use a textual entailment model to analyze small fragments of texts independently. Our second approach, an unstructured model, assumes that it is not sufficient to analyze different pieces of text independently. Instead, the complete text should be analyzed as a whole to identify desire fulfillment. Our third approach, a structured model, is based on the hypothesis that identifying desire fulfillment requires an understanding of the narrative structure and models the same using latent variables. We compare performances of these models on two different datasets that we have annotated and release. Our experiments establish the need to incorporate the narrative structure of the storyline offered by the text to better understand desire fulfillment.


\fontsize{9pt}{9.8pt} \selectfont
\bibliographystyle{abbrv}
\bibliography{main}

\end{document}